\documentclass[10pt,twocolumn,letterpaper]{article} 
\usepackage[pagenumbers]{wacv}

\usepackage{url}

\usepackage{xcolor}

\usepackage{array}
\usepackage{url}
\usepackage{booktabs}
\usepackage{graphicx}

\usepackage{multicol}
\usepackage{multirow}
\usepackage{pifont}
\begin{document}

\title{GaitFormer: Learning Gait Representations with Noisy Multi-Task Learning}

\author{Adrian Cosma\\
University Politehnica of Bucharest\\
Bucharest, Romania\\
{\tt\small cosma.i.adrian@gmail.com}
\and
Emilian Radoi\\
University Politehnica of Bucharest\\
Bucharest, Romania\\
{\tt\small emilian.radoi@upb.ro}
}

\maketitle

\begin{abstract}
Gait analysis is proven to be a reliable way to perform person identification without relying on subject cooperation. Walking is a biometric that does not significantly change in short periods of time and can be regarded as unique to each person. So far, the study of gait analysis focused mostly on identification and demographics estimation, without considering many of the pedestrian attributes that appearance-based methods rely on. In this work, alongside gait-based person identification, we explore pedestrian attribute identification solely from movement patterns. We propose DenseGait, the largest dataset for pretraining gait analysis systems containing 217K anonymized tracklets, annotated automatically with 42 appearance attributes. DenseGait is constructed by automatically processing video streams and offers the full array of gait covariates present in the real world. We make the dataset available to the research community. Additionally, we propose GaitFormer, a transformer-based model that after pretraining in a multi-task fashion on DenseGait, achieves 92.5\% accuracy on CASIA-B and 85.33\% on FVG, without utilizing any manually annotated data. This corresponds to a +14.2\% and +9.67\% accuracy increase compared to similar methods. Moreover, GaitFormer is able to accurately identify gender information and a multitude of appearance attributes utilizing only movement patterns. The code to reproduce the experiments is made publicly.

\end{abstract}

\section{Introduction}
Technologies relying on facial and pedestrian analysis play a crucial role in intelligent video surveillance and security systems. Facial and pedestrian analysis systems have become the norm in video intelligence, such systems being deployed ubiquitously. However, appearance-based pedestrian re-identification \cite{ye2021deep} and facial recognition models \cite{wang2021deep} invariably suffer from extrinsic factors related to camera viewpoint and resolution, and to the change in a person's appearance such as different clothing, hairstyles and accessories. Moreover, due to the proliferation of privacy laws such as GDPR, it is increasingly difficult to deploy appearance-based solutions for video-intelligence. Human movement is highly correlated with many internal and external aspects of a particular individual including age, gender, body mass index, clothing, carrying conditions, emotions and personality \cite{nixon2005human}. The manner of walking is unique to each person, it does not significantly change in short periods of time \cite{00003677-200304000-00009} and cannot be easily faked to impersonate another person \cite{kumar2021treadmill}. Gait analysis has gained significant attention in recent years \cite{singh2018vision,makihara2020gait}, due to solving many of the problems of appearance-based technologies without relying on the direct cooperation of subjects. However, compared to appearance-based methods, gait analysis is intrinsically harder to perform with reliable accuracy, due to the influence of many confounding factors that affect the manner of walking. This problem is tackled in literature in two major ways, either by building specialized neural architectures that are invariant to walking variations \cite{fvg,sepas2020view,thapar2018vgr}, or by creating large-scale and diverse datasets for training \cite{cosma2021wildgait,zhu2021gait,Makihara_CVATN2012,Xu_CVA2017,yu2006framework}.

One of the first attempts of building a large-scale gait recognition dataset is OU-ISIR \cite{Xu_CVA2017}, which is comprised of 10,307 identities that walk in a straight line for a short duration of time. Such a dataset is severely limited by its lack of walking variability, having only viewpoint change as a confounding factor. Building sufficiently large datasets that account for all the walking variations imply an immense annotation effort. For example, the GREW benchmark \cite{zhu2021gait} for gait-based identification, reportedly took 3 months of continuous manual annotation by 20 workers. In contrast, automatic, weakly annotated datasets are much easier to gather by leveraging existing state-of-the-art models---UWG \cite{cosma2021wildgait}, a comparatively large dataset of individual walking tracklets proved to be a promising new direction in the field. Increasing the dataset size is indeed correlated with performance on downstream gait recognition benchmarks \cite{cosma2021wildgait}, even though no manual annotations are provided. One limitation of these datasets is that they are annotated with attributes per individual only sparsely, and not addressing the problem of pedestrian attribute identification (PAI), currently performed only through appearance-based methods \cite{liu2017hydraplus,tang2019improving,jia2020rethinking}. Walking pedestrians are often annotated only with their gender, age, and camera viewpoint \cite{yu2006framework,Xu_CVA2017,fvg,zhu2021gait}. Even though gait-based demographic identification is a viable method for pedestrian \mbox{analysis \cite{catruna2021face}}, it is also severely limited by the lack of data.  Also, many attributes from PAI networks such as gender, age and body type have a definite impact on walking patterns \cite{KO20111974,ko2010age,choi2021body}, and we posit that they can be identified with a reasonable degree of accuracy using only movement patterns and not utilizing appearance information.

We propose DenseGait, the largest gait dataset for pretraining to date, containing 217k anonymized tracklets in the form of skeleton sequences, automatically gathered by processing real-world surveillance streams through state-of-the-art models for pose estimation and pose tracking. An ensemble of PAI networks was used to densely annotate each skeleton sequence with 42 appearance attributes such as their gender, age group, body fat, camera viewpoint, clothing information and apparent action. The purpose of DenseGait is to be used for pretraining networks for gait recognition and attribute identification, it is not suitable for evaluation since it is annotated automatically and does not contain manual, ground-truth labels. DenseGait contains walking individuals in real scenarios, it is markerless, non-treadmill, and avoids unnatural and constrictive laboratory conditions, which have been shown to affect gait \cite{takayanagi2019relationship}. It practically contains the full array of factors that are present in real world gait patterns. 

The dataset is fully anonymized, and any information pertaining to individual identities is removed, such as the time, location and source of the video stream, and the appearance and height information of the person. DenseGait is a gait analysis dataset primarily intended for pretraining neural models---using it to explicitly identify the individuals within it is highly unfeasible, requiring extensive external information about the individuals, such as personal identifying information (i.e., their name or ID) and a baseline gait pattern. According to GDPR\footnote{\url{https://eur-lex.europa.eu/eli/reg/2016/679/oj}, accessed on 1 July 2022)} legislation, data used for research purposes can be used if anonymized. Moreover, anonymized data does not conform to the rigors of personal data and can be processed without explicit consent. Nevertheless, \textit{any attempt to use of DenseGait to explicitly identify individuals present in it is highly discouraged.} %MDPI: Please confirm if the italics should be retained. Author: Yes, italics should be retained.

We chose to utilize only skeleton sequences for gait analysis, as current appearance-based methods that rely on silhouettes are not privacy preserving, potentially allowing for identification based only on the person's appearance, rather than their movement \cite{liu2004toward}.
Skeleton sequences encode only the movement of the person, abstracting away any visual queues regarding identity and attributes. Moreover, skeleton-based solutions have the potential to generalize across tasks such as action recognition, allowing for a flexible and extensible computation.

DenseGait, compared to other similar datasets \cite{cosma2021wildgait}, contains 10$\times$ more sequences and is automatically annotated with 42 appearance attributes through a pretrained PAI ensemble (Table \ref{tab:pai-network-attributes}). In total, 60 h of video streams were processed, having a cumulative walking duration of pedestrians of 410 h. We release the dataset under open credentialized access, for research purposes only, under CC-BY-NC-ND-4.0 \footnote{\url{https://creativecommons.org/licenses/by-nc-nd/4.0/legalcode}, accessed on 1 July 2022} License.

\begin{table*}[hbt!]
\small
    \caption{List of attributes extracted by each network in the PAI ensemble. Each network is trained on a different dataset, with a separate set of attributes. After coalescing similar attributes and eliminating appearance-only attributes, we obtain 42 appearance attributes.}
    \label{tab:pai-network-attributes}
    \begin{tabular}{p{0.26\textwidth}p{0.3\textwidth}p{0.35\textwidth}}
        \toprule
        \textbf{PA100k} & \textbf{PETA} & \textbf{RAP}\\
        \midrule
        Female,
        AgeOver60,
        Age18-60,
        AgeLess18,
        Front,
        Side,
        Back,
        Hat,
        Glasses,
        HandBag,
        ShoulderBag,
        Backpack,
        HoldObjectsInFront,
        ShortSleeve,
        LongSleeve,
        UpperStride,
        UpperLogo,
        UpperPlaid,
        UpperSplice,
        LowerStripe,
        LowerPattern,
        LongCoat,
        Trousers,
        Shorts,
        Skirt \& Dress,
        Boots
        & 
        Age16--30,
        Age31--45,
        Age46--60,
        AgeAbove61,
        Backpack,
        CarryingOther,
        Casual lower,
        Casual upper,
        Formal lower,
        Formal upper,
        Hat,
        Jacket,
        Jeans,
        LeatherShoes,
        Logo,
        LongHair,
        Male,
        Messenger Bag,
        Muffler,
        No accessory,
        No carrying,
        Plaid,
        PlasticBags,
        Sandals,
        Shoes,
        Shorts,
        Short Sleeve,
        Skirt,
        Sneaker,
        Stripes,
        Sunglasses,
        Trousers,
        TShirt,
        UpperOther,
        V-Neck
        & 
        Female,
        AgeLess16,
        Age17--30,
        Age31--45,
        BodyFat,
        BodyNormal,
        BodyThin,
        Customer,
        Clerk,
        BaldHead,
        LongHair,
        BlackHair,
        Hat,
        Glasses,
        Muffler,
        Shirt,
        Sweater,
        Vest,
        TShirt,
        Cotton,
        Jacket,
        Suit-Up,
        Tight,
        ShortSleeve,
        LongTrousers,
        Skirt,
        ShortSkirt,
        Dress,
        Jeans,
        TightTrousers,
        LeatherShoes,
        SportShoes,
        Boots,
        ClothShoes,
        CasualShoes,
        Backpack,
        SSBag,
        HandBag,
        Box,
        PlasticBags,
        PaperBag,
        HandTrunk,
        OtherAttchment,
        Calling,
        Talking,
        Gathering,
        Holding,
        Pusing,
        Pulling,
        CarryingbyArm,
        CarryingbyHand\\
        \bottomrule
        \end{tabular}
    \end{table*}

We also propose GaitFormer, a multi-task transformer-based architecture \cite{NIPS2017_3f5ee243} that is pretrained on DenseGait in a self-supervised manner, being able to perform exceptionally well in zero-shot gait recognition scenarios on benchmark datasets, achieving 92.5\% identification accuracy from direct transfer on the popular CASIA-B dataset, without using any manually annotated data. Moreover, it obtains good results on demographic and pedestrian attribute identification from walking patterns, with no manual annotations. GaitFormer represents the first use of a plain transformer encoder architecture in gait skeleton sequence processing, without relying on hand-crafted architectural modifications as in the case of graph neural networks \cite{yan2018spatial,plizzari2021spatial}.

This paper makes the following contributions:
\begin{enumerate}
    \item We release DenseGait, the largest dataset of skeleton walking sequences, densely annotated with appearance information, for use in pretraining neural architectures that can be further fine-tuned on specific gait analysis tasks.  The dataset can be found at \url{https://bit.ly/3SLO8RW}, under open credentialized access, for research purposes only.
    \item We propose GaitFormer, a multi-task transformer that is pretrained on the DenseGait dataset and achieves exceptional results in zero-shot gait recognition scenarios on benchmark datasets, achieving 92.52\% accuracy on CASIA-B and 85.33\% on FVG, without training on any manually annotated data (+14.2\% and +9.67\% increase compared to similar methods \cite{cosma2021wildgait}). The code is made publicly available at: \url{https://github.com/cosmaadrian/gaitformer}
    \item We explore the performance of GaitFormer on other gait analysis tasks, such as gait-based gender estimation and attribute identification. 
\end{enumerate}

\section{Related Work}
\subsection{Gait Analysis}
Video gait analysis encompasses research efforts dedicated to automatically estimate and predict various aspects of a walking person. Research has been mostly dedicated into gait-based person recognition, with many benchmark datasets \cite{fvg,cosma2021wildgait,hofmann2014tum,Xu_CVA2017,chen2020simple,shutler2004large,sarkar2005humanid,zhu2021gait} available for training and testing models. Moreover, there have been improvements in areas such as estimating demographics information \cite{catruna2021face,do2020real}, emotion detection \cite{xu2020emotion} and ethnicity estimation \cite{zhang2010ethnicity} from only movement patterns. \cite{sepas2022deep} proposed a taxonomy to organize the existing works in the field of gait recognition. In this work, we focus mainly on body representation, as we made a deliberate choice of providing DenseGait with only movement information for anonymization. Broadly, works in gait analysis can be divided into two major approaches in terms of body representation: silhouette-based and skeleton-based.

\subsubsection{Silhouette-based Solutions}
Silhouette-based approaches make use of silhouettes of walking individuals estimated either through background subtraction methods or through instance segmentation and tracking. Silhouettes are used in various forms, either in a condensed \mbox{representation \cite{gei-original,10.1007/978-3-642-15549-9_19,5522296}}, or as a sequences, as it is the norm in more modern methods \cite{chao2019gaitset,fan2020gaitpart,lin2021gait,fvg}. Most notably, \mbox{GaitSet \cite{chao2019gaitset}} processes the silhouettes as a set, as opposed to preserving the temporal information present in a sequence. As such, the authors can include silhouettes from multiple videos of the same walking subjects, achieving good invariance to walking variations. GaitPart \cite{fan2020gaitpart} processes the temporal variation of each individual body part separately in a Micro-motion Capture Module (MCM), taking inspiration from model-based approaches. Each body part exhibits different visual queues and temporal variation and the authors propose to combine the each feature part to construct the final gait representation. Recently, Lin et al. \cite{lin2021gait}, advance the construction of neural architectures for processing silhouette sequences by proposing a Global-Local Feature Extractor (GLFE), which obtains good results on benchmark datasets. Zhang et al. \cite{fvg} propose GaitNet, a model which directly makes use of the appearance of the individual and is able to output invariant feature representations for gait recognition. Moreover, they also propose FVG, a dataset with \mbox{226 individuals}, only from the front-view angle, one of the more challenging angles in gait analysis, due to the lack of perceived variation in limb movements.

\subsubsection{Skeleton-based Solutions}
Skeleton-based approaches, on the other hand, avoid making use of appearance information in the form of silhouettes, and instead focus on the moving anatomical skeleton of the person, effectively processing only movement patterns. Approaches typically imply processing walking sequences with a pose estimation \cite{li2018crowdpose} model, and processing the resulting skeletons with a neural network, either by adapting conventional CNN modules \cite{9324873}, or with an LSTM \cite{10.1007/978-3-319-69923-3_51,An2018ImprovingGR}. More modern approaches make use of graph neural networks to model the relationships between human joints \cite{li2020jointsgait,9721551}. Liao et al. \cite{10.1007/978-3-319-69923-3_51} make use of a combined CNN and LSTM architecture to model 2D skeleton sequences. A later improvement makes use of 3D skeletons \cite{An2018ImprovingGR} to further improve results. Li et al. \cite{li2020jointsgait} propose a graph-based convolutional architecture to process skeleton sequences, and a Joints Relationship Pyramid Mapping to map spatio-temporal gait features into a discriminative feature space. Li and Zhao \cite{9721551} propose CycleGait, a graph-based approach that incorporates multiple walking paces in the augmentation procedure and obtains robust results in gait recognition on CASIA-B. In contrast to these approaches, we opted to take a data-driven approach, instead of an algorithmic approach, and use a standard transformer architecture and pretrain it on a large amount of weakly-labelled data. Recently, \cite{cosma2021wildgait} proposed an approach called WildGait to skeleton-based gait recognition, in which they automatically mine surveillance streams and pretrain a ST-GCN \cite{yan2018spatial} model in a self-supervised manner. Through fine-tuning, good results are obtained in recognition on CASIA-B and FVG. Similarly to WildGait, we also process publicly available surveillance streams, but increase the DenseGait dataset size by an order of magnitude. Moreover, we densely annotate each skeleton sequence with 42 appearance attributes for use in zero-shot attribute identification scenarios.

However, model-based approaches still lag behind methods utilizing appearance (i.e., silhouettes). This is most likely due to the imperfect extraction of skeletons by modern pose estimators, which struggle to accurately detect fine-grained movements at a distance. Moreover, using appearance-based methods is fundamentally easier, since a single silhouette can contain identifying information about a subject. For instance Xu et al. \cite{xu2020gait} obtained reasonable results for gait recognition using a single silhouette, which cannot be considered gait, as no temporal movement is being processed at all. This implies that recognition is performed through ``shortcuts'' in the form of appearance features (i.e., body composition, height, haircut, side-profile etc). For this reason, a more privacy-aware approach is to process only movement patterns, which constitutes the motivation for releasing DenseGait with only anonymized skeleton sequences, and disregarding silhouettes.

\subsection{Transformers and Self-Supervised Learning}

In recent years, there has been a insurgence of research in the area of self-supervised learning, mostly due to the extremely high performance obtained in natural language processing with models such as BERT \cite{devlin-etal-2019-bert} and GPT \cite{NEURIPS2020_1457c0d6}. Self-supervised learning presumes training models using aspects of the data itself as a supervisory signal. While initial efforts in computer vision relied on creating artificial pretext tasks \cite{gidaris2018unsupervised,doersch2015unsupervised,wang2020self}, the field is moving towards contrastive-based approaches \cite{chen2020simple,zbontar2021barlow,caron2021emerging}. Methods such as SimCLR \cite{chen2020simple}, Barlow Twins \cite{zbontar2021barlow} and Dino \cite{caron2021emerging} obtaining almost similar performance to direct supervision. Moreover, the transformer has proven to be a flexible architecture, capable of handling a multitude of modalities such as text \cite{devlin-etal-2019-bert}, images \cite{dosovitskiy2020image}, video \cite{zhang2021token}, speech \cite{8462506}, and highly benefit from large-scale pretraining \cite{beal2022billion}. Taking inspiration from related efforts to process non-textual data with transformers \cite{dosovitskiy2020image}, we construct GaitFormer by processing flattened skeletons as input ``tokens''. In this manner, any human bias related to hand-crafted graph relationships between the body joints is eliminated. Moreover, as opposed to graph networks such as ST-GCN \cite{plizzari2021spatial}, training a similarly large transformer encoder make more efficient use of computational resources, significantly reducing training time.

\section{Method}
\subsection{Dataset Construction}
For building the DenseGait dataset, we made use of public video streams (e.g., street cams), and processed them with AlphaPose \cite{li2018crowdpose}, a modern, state-of-the-art multi-person pose estimation model. AlphaPose's raw output is comprised of skeletons with $(x,y,c)$ coordinates for each of the 18 joints of the COCO skeleton format Lin et al. \cite{lin2014microsoft}, corresponding to 2D coordinates in the image plane and a prediction confidence score. We performed intra-camera tracking for each skeleton with on SortOH \cite{nasseri2021simple}. SortOH is based on the SORT \cite{bewley2016simple} algorithm, which relies only on coordinate information and not on appearance information. As opposed to DeepSORT\cite{Wojke2017simple} which makes use of person re-identification models, SortOH is only using coordinates and bounding box size for faster computation time while having comparably similar performance. SortOH ensures that tracking is not significantly affected by occlusions.

To ensure that the skeleton sequences can be properly processed by a deep learning model, we performed extensive data cleaning. We have filtered low confidence skeletons by computing the average confidence of each of the 18 joints, and in each sequence, skeletons with an average confidence of less than 0.5 were removed. Furthermore, skeletons with feet confidence less than 0.4 were removed. This step guarantees that the feet are visible and confidently detected---leg movement is one of the most important signals for gait analysis. In our processing, we chose a period length $T$ of 48 frames, which corresponds to approximately 2 full gait cycles on average \cite{00004623-196446020-00009}. Surveillance streams do not have the same frame rate between them, which makes the sequences have different paces and durations. As such, we filtered short tracklets which have a duration of less than $\frac{T * fps}{24}$. We consider 24 FPS to be real-time video speed, and each video was processed according to its own frame rate. Moreover, skeletons are linearly interpolated such that the pace and duration is unified across video streams.

Similar to \cite{cosma2021wildgait}, we further normalized each skeleton by centering at the pelvis coordinates $(x_{pelvis}, y_{pelvis})$ and scaling vertically by the distance between the head and the hips ($y_{neck} - y_{pelvis}$) and horizontally by the distance between the shoulders ($(x_{R.shoulder} - x_{L.shoulder})$). This procedure is detailed in Equations (\ref{eq:eq1}) and (\ref{eq:eq2}). The normalization procedure aligns the skeleton sequences in a similar manner to the alignment step in face recognition pipelines \cite{xu2021searching}. This step eliminated the height and body type information about the subject, ensuring that the person cannot be directly identified. 
\begin{equation}
    x_{joint} = \frac{x_{joint} - x_{pelvis}}{|x_{R.shoulder} - x_{L.shoulder}|}
    \label{eq:eq1}
\end{equation}
\begin{equation}
    y_{joint} = \frac{y_{joint} - y_{pelvis}}{|y_{neck} - y_{pelvis}|}
    \label{eq:eq2}
\end{equation}

However, body type information should be preserved through the analysis of the walking patterns. Moreover, normalization obscures the human position in the frame, to prevent identification of the source video stream.

Finally, we filtered standing/non-walking skeletons in each sequence by computing the average movement speed of the legs, which is indicative of the action the person is performing. As such, if the average leg speed is less than 0.0015 and higher than 0.09, the sequence was removed. The thresholds were determined through manual inspection of the sequences. This eliminated both standing skeleton sequences as well as sequences with erratic leg movement, which is most probably due to poor pose estimation output in \mbox{that case}.

DenseGait is fully anonymized. Any information regarding the identity of particular individuals in the dataset is eliminated, including appearance information (by keeping only movement information in the form of skeleton sequences), height and body proportions (through normalization), and the time, location, and source of the video stream. Identifying individuals in DenseGait is highly unfeasible, as it requires external information (i.e., name, email, ID, etc.) and specific collection of gait patterns.

The final dataset contains 217k anonymized tracklets, with a combined length of 410 h. DenseGait is currently the largest dataset of skeleton sequences for use in pretraining gait analysis models. Table \ref{tab:dataset-comparison} showcases a comparison between DenseGait and other popular gait recognition datasets. Since the skeleton sequences are collected automatically through pose tracking, it is impossible to quantify exactly the number of different identities in the dataset, as, in some cases, tracking might be lost due to occlusions. However, DenseGait contains a significantly larger number of tracklets compared to other available datasets while also being automatically densely annotated with 42 appearance attributes. In the case of UWG \cite{cosma2021wildgait} and DenseGait, the datasets do not contain explicit covariates for each identity, but rather covariates in terms of viewing angle, carrying conditions, clothing change, and apparent action are present across the tracklet duration, similar to GREW \cite{zhu2021gait}.

\begin{figure}[hbt!]
    \centering
    \includegraphics[width=0.7\linewidth]{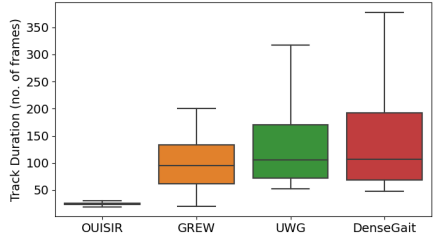}
    \caption{Comparison between existing large-scale skeleton gait databases and DenseGait in terms of distributions of tracklet duration. DenseGait is an order of magnitude larger than the next largest skeleton database, while having a longer average duration (136 frames UWG vs 162 frames DenseGait).}
    \label{fig:track-durations}
\end{figure}

Similarly to UWG \cite{cosma2021wildgait}, DenseGait does not contain multiple walks per person, rather each tracklet is considered a unique identity. Compared to other large-scale datasets, DenseGait tracks individuals for a longer duration, which makes it suitable for use in self-supervised pretraining, as longer tracked walking usually contains more variability for a single person. Figure \ref{fig:track-durations} shows boxplots with a five-number summary descriptive statistics for the distribution of track durations in each dataset. DenseGait has a mean tracklet duration of 162 frames, which is significantly larger (z-test \emph{p} $<$ 0.0001) compared to other datasets: CASIA-B \cite{yu2006framework}---83 frames, FVG \cite{fvg}---97 frames, GREW \cite{zhu2021gait}---98 frames, UWG \cite{cosma2021wildgait}---136 frames). Due to potential loss of tracking information, the dataset is noisy, and can be used only for self-supervised pretraining. 
%{\color{red} Pedestrians processed in DenseGait are extracted from surveillance streams across the world, from all major continents and from all ages, genders and ethnicities, to limit as much as possible any form of dataset bias \cite{zhang2010ethnicity}}.

\begin{table*}[hbt!]
\setlength{\tabcolsep}{4mm}
        \small
       \centering
        \caption{Comparison of popular datasets for gait recognition. DenseGait is an order of magnitude larger, has more identities in terms of skeleton sequences (highlighted in \textbf{bold}), and each sequence is annotated with 42 appearance attributes. * Approximate number given by pose tracker. $\dagger$ Implicit covariates across tracking duration.}
        \label{tab:dataset-comparison}
        \resizebox{\linewidth}{!}{
        \begin{tabular}{lccccc}
        \toprule
        \textbf{Dataset} & \textbf{\# IDs} & \textbf{Sequences} & \textbf{Covariates} & \textbf{Views} & \textbf{Env.} \\
             \midrule
        USF HumanID \cite{sarkar2005humanid} & 122 & 1870 & Y & 2 & Outdoor \\
        TUM-GAID \cite{hofmann2014tum} & 305 & 3370 & Y & 1 & Outdoor \\
        FVG \cite{fvg} & 226 & 2857 & Y & 1 & Outdoor \\
        CASIA-B \cite{yu2006framework} & 124 & 13,640 & Y & 11 & Indoor  \\
        OU-ISIR \cite{Xu_CVA2017} & 10,307 & 144,298 & N & 14 & Indoor \\
        GREW \cite{zhu2021gait} & 26,000 & 128,000 & Y & - & Outdoor\\
        UWG \cite{cosma2021wildgait} & 38,502 * & 38,502 & Y $^\dagger$ & - & Outdoor\\
        \midrule
        DenseGait (\textbf{ours}) & \textbf{217,954} * & \textbf{217,954} & Y $^\dagger$ & - & Outdoor\\
        \bottomrule
        \end{tabular}
        }
    \end{table*}

\subsection{Annotations with Appearance Attributes}

\begin{figure*}[hbt!]
    \centering
    \includegraphics[width=0.97\linewidth]{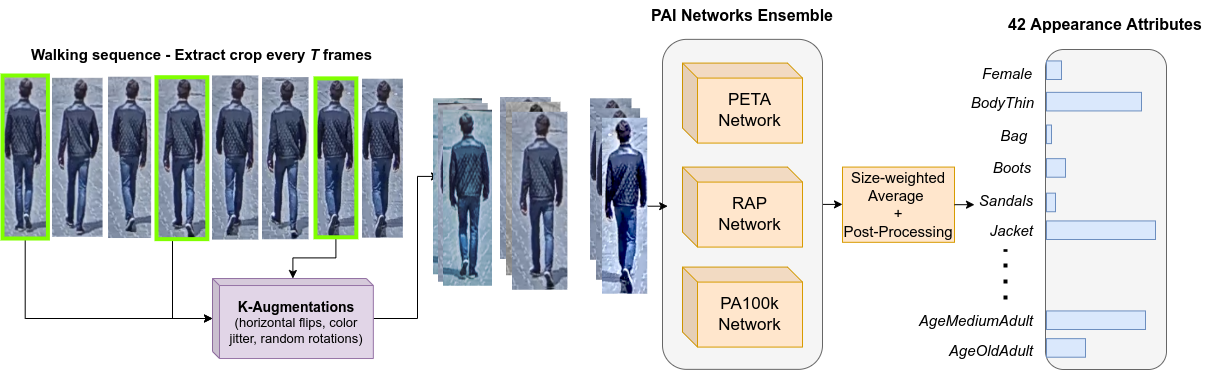}
    \caption{Overview of the automatic annotation procedure for the 42 appearance attributes. To robustly annotate attributes, an ensemble of pretrained networks is used in conjunction with multiple augmentations of the same crop. Predictions across the sequence are averaged according to their bounding-box area.}
    \label{fig:uwg2-attributes}
\end{figure*}

Appearance attributes are essential for pretraining for tasks such as gender \mbox{estimation \cite{catruna2021face}}, age estimation \cite{Li2018} and pedestrian attribute identification \cite{liu2017hydraplus,tang2019improving,jia2020rethinking}. To ensure that the dataset is densely annotated with appearance attributes, we made use of an ensemble of pretrained PAI networks, each trained on different popular PAI datasets. Specifically, we employed three InceptionV3 \cite{szegedy2016rethinking} networks trained on RAP \cite{li2019richly}, PETA \cite{deng2014pedestrian} and \mbox{PA100k \cite{liu2017hydraplus}}, respectively. Figure \ref{fig:uwg2-attributes} showcases the annotation procedure.

Since each dataset has a different set of pedestrian attributes, we averaged similar classes (e.g., \textit{AgeLess16} and \textit{AgeLess18} into \textit{AgeChild}), coalesced similar classes (e.g., \textit{Formal} and \textit{Suit-Up} into \textit{FormalWear}) and removed attributes that cannot evidently be estimated from movement patterns (e.g., \textit{BaldHead}, \textit{Hat}, \textit{V-Neck}, \textit{Glasses}, \textit{Plaid} etc.). 

For a particular sequence, we take the cropped image of the pedestrian at every $T$ frames (where $T$ is the period length), and randomly augment it $k = 4$ times (e.g., random horizontal flips, color jitter and small random rotation). For each crop, each augmented version is then processed by a PAI network and the results are averaged such that the output is robust to noise \cite{sohn2020fixmatch}. Finally, to have a unified prediction for the walking sequence, results are averaged according to the size of the bounding box relative to the image, similar to Catruna et al. \cite{catruna2021face}. Predictions on larger crops have a higher weight, with the assumption that the pedestrian appearance is more clearly distinguishable when closer to the camera.

\begin{figure*}[hbt!]
    \centering
    \includegraphics[width=0.9\linewidth]{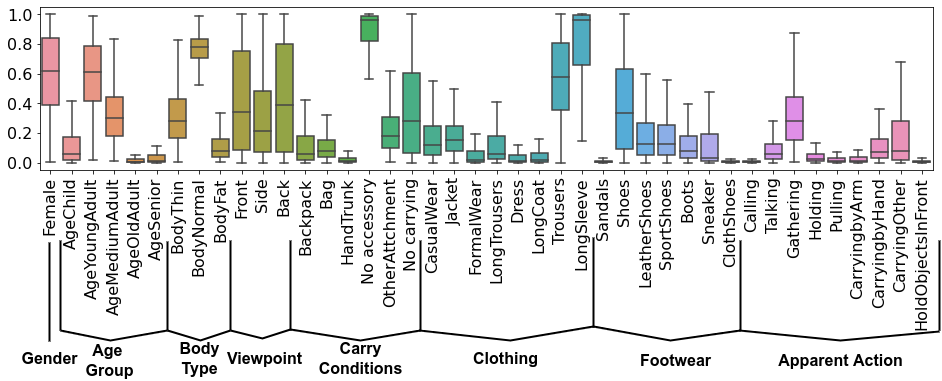}
    \caption{Distribution of the 42 appearance attributes in DenseGait. The dataset is annotated in a fine-grained manner with attributes ranging from internal aspects of the person (Gender, Age Group, Body Type) to appearance only labels (Clothing, Footwear).} %MDPI: 1. please confirm if italic is necessary? Author: Not neccessary, removed. 2. please confirm if different colors need explanations in image? Author: No extra explanations needed, each color is a different attribute, and the colors are the same with Figure 8.
    \label{fig:attribute-distribution}
\end{figure*}

Figure \ref{fig:attribute-distribution} showcases the final list of attributes, and their distribution across the dataset. We have a total of 42 attributes, split into 8 groups: \textit{Gender}, \textit{Age Group}, \textit{Body Type}, \textit{Viewpoint}, \textit{Carry Conditions}, \textit{Clothing}, \textit{Footwear} and \textit{Apparent Action}. For the final annotations, we chose to keep the soft-labels and not round them, as utilizing soft-labels for model training was shown to be a more robust approach when dealing with noisy data \cite{sohn2020fixmatch}.

\begin{figure}[hbt!]
    \centering
    \includegraphics[width=0.8\linewidth]{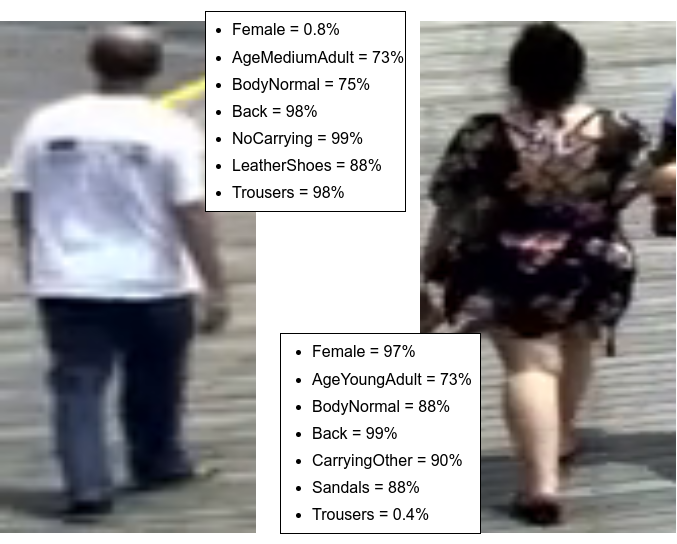}
    \caption{Qualitative examples for selected attributes from the PAI ensemble. The networks correctly identify gender, age group and viewpoint. However, in some cases, clothing and, more specifically, footwear are more difficult to estimate in low resolution scenarios.}
    \label{fig:attributes-qualitative}
\end{figure}

Figure \ref{fig:attributes-qualitative} showcases selected examples of attribute predictions from the PAI ensemble. Since surveillance cameras usually have low resolution and the subject might be far away from the camera, some pedestrian crops are blurry and might affect prediction by the PAI ensemble. For gender, age group, body composition and viewpoint, the models are confidently identifying these attributes. However, for specific pieces of clothing (i.e., footwear: Sandals/LeatherShoes), predictions are not always reliable, due to the low resolution of some of the crops, but the errors are negligible when taking into account the scale of the dataset.

\subsection{Description of Model Architecture}

For pretraining on the DenseGait dataset for the tasks of gait-based recognition and attribute identification, we chose to adapt the popular transformer encoder \mbox{architecture \cite{NIPS2017_3f5ee243}} to handle skeleton sequences. Initially, transformers were immensely successful in handling sequential data in the form of text, effectively replacing LSTM \cite{hochreiter1997long} networks, the de facto approach for these problems. However, lately, transformers have been used in a variety of problems, being able to handle images \cite{dosovitskiy2020image}, video \cite{zhang2021token} and multi-modal data \cite{gabeur2020multi}. Moreover, transformer architectures in particular highly benefit from large-scale, self-supervised pretraining \cite{caron2021emerging,devlin-etal-2019-bert,NEURIPS2020_1457c0d6}, allowing models to be effectively fine-tuned on more specific datasets with small amounts of annotated data.

\begin{figure*}[hbt!]
    \centering
    \includegraphics[width=0.8\linewidth]{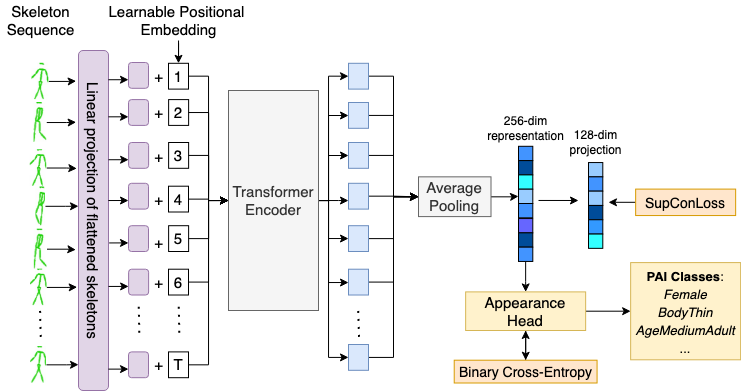}
    \caption{Overview of GaitFormer (Multi-Task) training procedure. Flattened skeletons are linearly projected using a standard feed-forward layer and fed into a transformer encoder. The vectorized representations are average pooled and the resulting 256-dimensional vector is used for estimating the identity and to estimate the 42 appearance attributes through the ``Appearance Head''. The contrastive objective (SupConLoss) is applied to a lower 128-dimensional linear projection, similar to the approach in SimCLR {\cite{chen2020simple}} 
.}
    \label{fig:model-training}
\end{figure*}

To handle skeleton sequences, we abstain from making any hand-crafted architectural modifications, as in the case of Plizzari et al. \cite{plizzari2021spatial}, which uses a hybrid approach by combining graph computation on the skeleton and using multi-head attention on the extracted features. Instead, we take inspiration from ViT \cite{dosovitskiy2020image}, which processes images as a sequence of flattened patches that are fed into a standard transformer encoder network. Figure \ref{fig:model-training} showcases the training procedure for GaitFormer in the multi-task training regime. Each skeleton is flattened into a 54 dimensional vector and is linearly projected with a standard learnable feed-forward layer into a 256 dimensional space. Each skeleton projection is then fed into a transformer encoder network. We opted for learnable positional embedding that is added to each projection instead of concatenated, to avoid increasing the dimensionality. After the transformer encoder, representations for each skeleton are averaged, and a final linear feed-forward layer of 256 elements is used as the final embedding. Further, as described in SimCLR \cite{chen2020simple}, we used an additional 128-dimension linear layer for training with a supervised contrastive objective \cite{NEURIPS2020_d89a66c7}. Additionally, a linear layer is used as appearance head to estimate the pedestrian attributes that is trained using a standard binary-crossentropy loss.

We used three different model sizes for the transformer encoder in our experiments, with 4 encoder layers (\textit{SM}), 8 encoder layers (\textit{MD}) and 12 encoder layers (\textit{XL}). In all types of architectures, 8 attention heads were used, and the internal feed-forward dimensionality was 256 \cite{NIPS2017_3f5ee243}.

\subsection{Training Details}

For training on DenseGait, we chose to use contrastive learning \cite{chen2020simple} as a supervisory signal. By design, contrastive methods work by attracting representations belonging to the same class, while simultaneously repelling samples from different classes. This paradigm is identical to the objective for recognition problems, which constitues one of the main tasks in gait analysis.
Specifically, we used SupConLoss \cite{NEURIPS2020_d89a66c7}, with a temperature of $\tau = 0.001$, alongside a two-view sampler for each skeleton in the batch. SupConLoss assumes a multi-viewed batch, with multiple augmentations for the same sample. Each view of a skeleton sequence is randomly augmented by the standard suite of augmentations for this data modality: random sequence crops of fixed length of $T = 48$, random flips with 50\% probability, random paces \cite{wang2020self}, and random gaussian noise added to joints coordinates. Let $i\in I\equiv\{1 \dots 2N\}$ be the index of an arbitrary augmented sample. SupConLoss is defined as:
\begin{equation}
    \mathcal{L}^{sup} = \sum_{i \in I} \frac{-1}{|P(i)|}\sum_{p\in P(i)}\log\frac{\exp(z_i \cdot z_p/\tau)}{\sum_{a\in A(i)} \exp(z_i \cdot z_a/\tau)}
    \label{eq:supconloss}
\end{equation}

In Equation (\ref{eq:supconloss}), $z_l = Enc(\widetilde{x_l})$ denotes the embedding of a skeleton sequence $x_l$, ``$\cdot$'' denotes the dot product operation and $A(i) \equiv I \setminus \{i\}$. Moreover, $P(i) \equiv \{p \in A(i): \widetilde{y}_p = \widetilde{y}_i\}$ is the set of indices of all positives in the multi-viewed batch distinct from $i$. In our case, the positive pairs are constructed by two different augmentations of the same skeleton sequence. The variability of the two augmentations is higher if the skeleton is tracked for a longer duration of time, as the walking individual might change direction.

As suggested in Chen et al. \cite{chen2020simple}, the supervisory signal given by SupConLoss is applied to a lower dimensional embedding (128 dimensions) to avoid the curse of dimensionality.

For predicting appearance attributes, which is a multi-label problem, we used a standard binary-crossentropy loss between each appearance label ($p_i$) and its corresponding prediction ($y_i$) (Equation (\ref{eq:appearance-loss})). As previously mentioned, we keep the soft labels as a supervisory signal, to prevent the network from overfitting and be more robust to noisy or incorrect labels \cite{muller2019does}. Moreover, since learning appearance labels can regarded as a knowledge distillation problem between the PAI ensemble and the transformer network, soft labels help improve the distillation process \cite{44873}.
\begin{equation}
    \mathcal{L}_{appearance} = -{(y_i\log(p_i) + (1 - y_i)\log(1 - p_i)))}
    \label{eq:appearance-loss}
\end{equation}

In multi-task (MT) training scenarios, we used a combination of the two losses, with a weight penalty of $\lambda = 0.5$ on the appearance loss $\mathcal{L}_{appearance}$. We chose $\lambda = 0.5$ empirically, such that the two losses have similar magnitudes. The final loss function is defined as:
\begin{equation}
    \mathcal{L}_{final} = \mathcal{L}_{SupCon} + \lambda \mathcal{L}_{appearance}
\end{equation}

In plain contrastive training scenarios, we employ only the SupConLoss, without predicting attributes (i.e., $\mathcal{L}_{final} = \mathcal{L}_{SupCon}$).

The motivation for pre-training the network in a multi-task setting is that the network not only learns to cluster walking sequences by their identity, but also to take appearance attributes into account. For instance, predicting the gender and age, even if they are not completely reliable, could prove useful for gait recognition, as demographics can be considered soft-biometrics, allowing the network to automatically filter identities by these attributes. On the other hand, in contrastive-only scenario, the network is under a classical self-supervised regime.

We used a batch size of 1024 across our experiments, with a cyclical learning rate \cite{smith2017cyclical} ranging from 0.0001 and 0.001 across 20 epochs. We trained all models for 400 epochs.

\section{Experiments and Results}
This section explores the performance of GaitFormer on gait-based recognition, gender identification and pedestrian attribute identification. We are primarily interested in evaluating the model in scenarios with low amounts of annotated data and we opted to use the two popular benchmark datasets originally constructed for gait recognition: \mbox{CASIA-B \cite{yu2006framework}} and FVG \cite{fvg}. For gender estimation, we manually annotated the gender information for each identity in the two datasets and constructed CASIA-gender and FVG-gender. We briefly describe each dataset below.

We chose CASIA-B to compare with other skeleton-based gait recognition models, since it is one of the most popular gait recognition datasets in literature. It contains \mbox{124 subjects} walking indoors in a straight line, captured with 11 synchronized cameras with three walking variations---normal walking (NM), clothing change (CL) and carry conditions (BG). According to Yu et al. \cite{yu2006framework}, the first 62 subjects are used for training and the rest for evaluation. CASIA-gender consists of manually annotated the subjects in CASIA-B with gender information, having a split of 92 males and 32 females. We maintain the training and validation splits from the recognition task, using the first 62 subjects for training (\mbox{44 males} and 18 females) and the rest for validation (48 males and 14 females). We use FVG to evaluate the robustness of GaitFormer, as it contains different covariates than CASIA-B such as varying degrees of walking speed, the passage of time and cluttered background. Moreover, FVG only contains walks from the front-view angle, which is more difficult for gait processing due to lower perceived limb variation. According to \mbox{Zhang et al. \cite{fvg}}, from the 226 identities present in FVG, the first 136 are used for training and the rest for testing. Similarly, {FVG-gender contains manual annotations with gender information, obtaining 149 males and 77 females. We maintain the training and validation splits from the recognition task, utilizing the first 136 individuals for training (83 males and 53 females) and the rest for validation (66 males and 24 females).

\subsection{Recognition}

We initially trained GaitFormer under two regimes: (i) contrastive only and (ii) multi-task (MT), which implies training with SupConLoss \cite{NEURIPS2020_d89a66c7} on the tracklet ID while simultaneously estimating the appearance attributes (Figure \ref{fig:model-training}). We experiment with three models sizes: SM---4 encoder layers (2.24M parameters), MD---8 encoder layers (4.35M parameters) and XL---12 encoder layers (6.46M parameters).

We pretrain GaitFormer on the DenseGait dataset under the mentioned conditions and directly evaluate recognition performance in terms of accuracy on CASIA-B and FVG, without fine-tuning. In all experiments we perform a deterministic crop in the middle of the skeleton sequences of T = 48 frames, and use no test-time augmentations. For each cropped skeleton sequence, features are extracted using the 256-dimensional representation and are normalized with the l$_2$ norm.  In Table \ref{tab:size-combined} we present results on the walking variations for each model size and training regime. For CASIA-B, we show mean accuracy where the gallery set contains all viewpoints except the probe angle, in the three evaluation scenarios: normal walking (NM), change in clothing (CL) and carry bag (CB). For FVG, we show accuracy results based on the evaluation protocols mentioned by Zhang et al. \cite{fvg}, corresponding to different walking scenarios (walk speed (WS), change in clothing (CL), carrying bag (CB), cluttered background (CBG) and ALL). Results show that unsupervised pretraining on DenseGait is a viable way to perform gait recognition, achieving an accuracy of 92.52\% on CASIA-B and 85.33\% on FVG, without any manually annotated data available. Notably, multi-task learning on appearance attributes provides a consistent positive gap in the downstream performance. 

\begin{table*}[hbt!]
        \small
\setlength{\tabcolsep}{2.97mm}
      \caption{GaitFormer direct transfer performance on gait recognition on CASIA-B and FVG datasets. We highlight in \textbf{bold} the best overall result for each dataset.}
       \label{tab:size-combined}
      \resizebox{\linewidth}{!}{
    \begin{tabular}{ccllllllll}
    \toprule
            & & \multicolumn{3}{c}{\textbf{CASIA-B}} &  \multicolumn{5}{c}{\textbf{FVG}} \\
            \midrule
           \textbf{Size} & \textbf{Training} & \textbf{NM} & \textbf{CL} & \textbf{CB} & \textbf{WS} &     \textbf{CB} &     \textbf{CL} &    \textbf{CBG} &    \textbf{ALL}  \\
           \midrule
          SM & Contrastive & 89.00 & 22.36 & 61.88 &  77.33 &  81.82 &  54.27 &  86.75 &  77.33 \\
        MD & Contrastive & 90.18 & 23.46 & 60.78 &  78.33 &  72.73 &  49.15 &  83.33 &  78.33 \\
       XL & Contrastive & 91.79 & 21.11 & 63.12 &  76.33 &  69.70 &  48.29 &  87.61 &  76.33  \\  
          \midrule
          SM & MT & 92.52 &  22.73 &  \textbf{67.16} & 84.67 &  81.82 & \textbf{59.40} &  \textbf{91.45} &  84.67  \\
        MD & MT &  \textbf{92.52} &  \textbf{23.31} & 65.10 &  \textbf{85.33} &  \textbf{87.88} &  53.42 &  88.89 &  \textbf{85.33} \\
       XL & MT & 90.69 &  20.75 &  60.34 &  85.00 &  81.82 &  51.71 &  91.03 &  85.00 \\
        \bottomrule 
    \end{tabular}
    }
    \end{table*}

Model size in terms of number of layers does not seem to considerably affect performance on benchmark datasets. GaitFormerMD (8 layers) fairs consistently better than GaitFormerXL (12 layers), while being similarly close to GaitFormerSM (4 layers). 

\begin{figure*}[hbt!]
    \centering
    \includegraphics[width=1.0\linewidth]{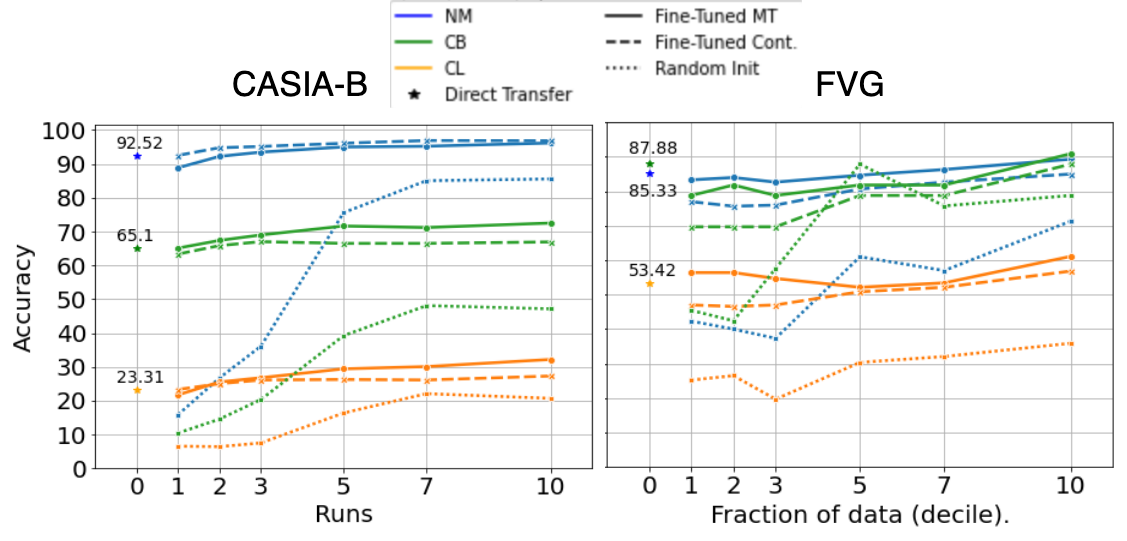}
    \caption{Fine-tuning results on gait recognition on CASIA-B and FVG, on progressively larger number of runs per identity. Compared to the same network randomly initialized, pretraining on DenseGait offers substantial improvements, even in the direct transfer regime. A consistent performance increase is obtained when also estimating attributes.}
    \label{fig:fine-tunes}
\end{figure*}

Figure \ref{fig:fine-tunes} compares GaitFormerMD pretrained on DenseGait in the two training regimes (contrastive only---Cont. and Multi-Task---MT) and GaitFormerMD randomly initialized. The networks were fine-tuned on progressively larger samples of the corresponding datasets: for CASIA-B, we sampled multiple runs for the same identity (from 1 to 10 runs per ID), and for FVG, we randomly sampled a percentage of runs per each identity. Models were fine-tuned using Layer-wise Learning Rate Decay (LLRD) \cite{zhang2020revisiting}, which implies a higher learning rate for top layers and a progressively lower learning rate for bottom layers. The learning rate was decreased linearly from 0.0001 to 0, across 200 epochs. The results show that unsupervised pretraining has a substantial effect on downstream performance especially in low data scenarios (direct transfer and 10\% of available data). Moreover, pretraining the model in Multi-Task learning regime, in which the network was tasked to estimate appearance attributes from movement alongside with the identity, provides a consistent increase in performance. 

Table \ref{tab:casia-sota} presents state-of-the-art results compared with other skeleton-based gait recognition models. We showcase the results of GaitFormerSM trained in the Multi-Task (MT) regime, without fine-tuning (direct) and tuned with all the available training data in CASIA-B. For comparison, we include WildGait \cite{cosma2021wildgait} with and without fine-tuning, as this model is also pretrained on a large dataset of skeleton sequences. We also compare with our implementation of GaitGraph Teepe et al. \cite{teepe2021gaitgraph}---a multi-branch ST-GCN which processes joint coordinates, velocities and bone angles, achieving great results on CASIA-B---and with a ST-GCN pretrained on DenseGait.

It is clear that the fine-tuned GaitFormerSM has very good results even without fine-tuning, achieving comparable results with the state of the art. Fine-tuning marginally increases the performance, achieving 96.2\% accuracy on normal walking (NM) and 72.5\% performance in carry bag (CB).

\begin{table*}[hbt!]
\setlength{\tabcolsep}{2.35mm}
\small
    \caption{GaitFormer comparison to other skeleton-based gait recognition methods on CASIA-B dataset. In all methods the gallery set contains all viewpoints except the proble angle. In \textbf{bold} and \underline{underline} we highlight the best and second best results for a particular viewpoint and walking condition.}\label{tab:casia-sota}
    \centering
        \resizebox{\linewidth}{!}{
        \begin{tabular}{p{2mm} l  rrrrrrrrrrrr}
        \toprule
             & \textbf{Method} &  \textbf{0\boldmath{$^{\circ}$}}  &    \textbf{18\boldmath{$^{\circ}$}}  &    \textbf{36\boldmath{$^{\circ}$}}  &    \textbf{54\boldmath{$^{\circ}$}}  &    \textbf{72\boldmath{$^{\circ}$}} & \textbf{90\boldmath{$^{\circ}$}}  &  \textbf{108\boldmath{$^{\circ}$}} &    \textbf{126\boldmath{$^{\circ}$}} &  \textbf{144\boldmath{$^{\circ}$}} &    \textbf{162$^{\circ}$} & \textbf{180\boldmath{$^{\circ}$}} & \textbf{Mean} \\
            \midrule
            \multirow{7}{*}{\textbf{NM}} &  GaitGraph & 79.8 &    89.5 &    91.1 &    92.7 &    87.9 &    89.5 &     94.35 &     95.1 &     92.7 &     93.5 &     80.6 & 89.7 \\
            & ST-GCN (DenseGait) & 89.5 &  89.5 &  95.1 &  87.9 &  81.4 &  68.5 &  64.5 &  89.5 &  88.7 &  84.6 &  82.2 &  83.8 \\
             & WildGait---direct & {72.6} &  {84.6} &  {90.3} &  {83.8} &  {63.7} &  {62.9} &  66.1 &  {83.0} &  {86.3} &  {84.6} &  {83.0} &  {78.3} \\
             & PoseFrame & 66.9 & 90.3 & 91.1 & 55.6 & 89.5 & \textbf{97.6} & \textbf{98.4} & 97.6 & 89.5 & 69.4 & 68.5 & 83.1 \\
             & GaitFormerSM---MT---direct & {\underline{94.3}} & \underline{97.5} &	\underline{99.2} &	\underline{98.4} &	79.8 &	80.6 &	89.5 &	\textbf{100.0} &	94.3 &	\underline{95.1}  &	\underline{88.7} &	92.5 \\
             & WildGait---tuned & {86.3} &    {96.0} &    {97.6} &    {94.3} &    \textbf{92.7} &    \underline{94.3} &     {94.3} &     {98.4} & \underline{97.6} &     {91.1} & {83.8} & \underline{93.4} \\
             & GaitFormerSM---MT---tuned & \textbf{96.7}	& \textbf{99.2} &	\textbf{100.0} &	\textbf{99.2} &	\underline{91.9} &	91.9 &	\underline{95.1} &	\underline{98.4} &	\textbf{96.7} &	\textbf{97.6} &	\textbf{91.1} &	\textbf{96.2} \\
            \midrule
            \multirow{7}{*}{\textbf{CL}} &  GaitGraph & 27.4 &    33.0 &    40.3 &    37.1 &    33.8 &    33.0 &     35.4 &     33.8 &     34.6 &     21.7 &     17.7 & 31.6\\
            &  ST-GCN (DenseGait) & 18.5 &  22.5 &  25.0 &  21.7 &  13.7 &  18.5 &  21.7 &  31.4 &  21.7 &  21.7 &  16.9 &  21.2\\
            & WildGait---direct & 12.1 & \underline{33.0} & {25.8} & 18.5 & {12.9} & 11.3 & {21.7} & {24.2} & {20.1} & {26.6} & {16.1} & {20.2} \\
            & PoseFrame & 13.7 & 29.0 & 20.2 & 19.4 & {28.2} & \textbf{53.2} & \textbf{57.3} & \textbf{52.4} & 25.8 & 26.6 & 21.0 & 31.5 \\
            & GaitFormerSM/MT---direct & 12.9 &	21.7 &	29.0 &	25.8 &	16.1 &	18.5 &	22.5 &	29.0 &	27.4 &	26.6 &	20.1 &	22.7 \\
            & WildGait---tuned & \underline{29.0} &    {32.2} &    \textbf{35.5} &    \textbf{40.3} &    \underline{26.6} &    25.0 &     38.7 &     38.7 &     31.4 &     \underline{34.6} &     \textbf{31.4} & \textbf{33.0}\\
            & GaitFormerSM/MT---tuned & \textbf{35.5} &	\textbf{35.5} &	\underline{33.8} &	\underline{33.8} &	20.9 &	30.6 &	31.4 &	31.4 &	\underline{28.2} &	\textbf{42.7} &	\underline{29.8} &	\underline{32.2} \\
            \midrule
            \multirow{7}{*}{\textbf{BG}} &  GaitGraph & 64.5 &    69.3 &    70.1 &    62.9 &    61.2 &    58.8 &     59.6 &     58.0 &     57.2 &     55.6 &     45.9 & 60.3 \\
            & ST-GCN (DenseGait) & 78.2 &  68.5 &  71.7 &  60.4 &  59.6 &  45.9 &  46.7 &  58.0 &  58.0 &  58.0 &  51.6 &  59.7 \\
            & WildGait---direct & {67.7} & {60.5} & {63.7} & {51.6} & {47.6} & 39.5 & 41.1 & {50.0} & {52.4} & {51.6} & {42.7} & {51.7} \\
            & PoseFrame  & 45.2 & 66.1 & 60.5 & 42.7 & \underline{58.1} & \textbf{84.7} & \textbf{79.8} & \textbf{82.3} & \underline{65.3} & 54.0 & 50.0 & 62.6 \\
            & GaitFormerSM/MT---direct & \underline{78.2} &	\underline{71.7} &	\textbf{84.7} &	\textbf{74.2} &	56.4 &	50.0 &	57.2 &	66.1 &	69.3 &	\underline{70.9} &	\underline{59.6} &	\underline{67.1} \\
            & WildGait---fine-tuned & {66.1} &    {70.1} &    {72.6} &    {65.3} &    56.4 &    64.5 &     65.3 &     67.7 &     57.2 &     {66.1} &     {52.4} & {64.0} \\
            & GaitFormerSM/MT---tuned & \textbf{82.2} &	\textbf{80.6} &	\underline{83.8} &	\underline{72.6} &	\textbf{62.9} &	\underline{69.3} &	\underline{68.5} &	\underline{70.1} &	\textbf{69.3} &	\textbf{77.4} &	\textbf{60.4} &	\textbf{72.5} \\\bottomrule
        \end{tabular}
        }
\end{table*}

\subsection{Comparison with ST-GCN and Other Pretraining Datasets}

In Table \ref{tab:dataset-architectures}, we compare GaitFormer with ST-GCN \cite{yan2018spatial} under different pretraining datasets. Reported results are mean accuracy across all angles for CASIA-B, under normal walking (NM) scenario, and accuracy under ALL scenario for FVG. The networks were not fine-tuned on these datasets; we present direct transfer performance after pretraining. We chose to pretrain on OU-ISIR \cite{Makihara_CVATN2012}, as this dataset is one of the most popular, large-scale datasets for gait recognition. However, OU-ISIR lacks data diversity, as all individuals are walking on a treadmill for a short duration, which is not the case for DenseGait. We also chose to pretrain on GREW \cite{zhu2021gait}, as it is also a diverse dataset collected in the wild, but contains fewer identities that walk for a comparably shorter duration of time.

Results show that, as a pretraining dataset, DenseGait is consistently outperforming GREW and OU-ISIR across the two architectures. These results are consistent with the insights in Figure \ref{fig:track-durations}, in which we posit that longer tracking duration for the individuals imply larger data diversity when pretraining in a contrastive self-supervised fashion, which directly improves performance.

 \begin{table}[hbt!]
        \small
        \caption{Comparison between GaitFormer and ST-GCN pretrained with Supervised Contrastive on GREW {\cite{zhu2021gait}}, OU-ISIR {\cite{Makihara_CVATN2012}} and our proposed DenseGait. Performance is directly correlated with mean tracklet duration on each dataset as shown in Figure \ref{fig:track-durations}.. We highlight in \textbf{bold} the best results for each architecture and dataset.}
        \label{tab:dataset-architectures}
        \resizebox{\linewidth}{!}{
       \begin{tabular}{llcc}
       \toprule
    \textbf{Backbone} & \textbf{Pretraining Data} & \textbf{CASIA-B (NM)} & \textbf{FVG (ALL)}\\
    \midrule
    \multirow{3}{*}{ST-GCN} 
    & OU-ISIR & 55.65 & 63.33\\
    & GREW & 61.14 & 56.67\\
    & DenseGait \textbf{(ours)} &  \textbf{83.80} & \textbf{75.28} \\
    \midrule
    \multirow{3}{*}{GaitFormer \textbf{(ours)}} 
    & OU-ISIR & 25.73 & 51.34\\
    & GREW & 65.40 & 64.04\\
    & DenseGait \textbf{(ours)} & \textbf{89.0}  & \textbf{77.33}\\
    \bottomrule
    \end{tabular}
    }
    \end{table}

\subsection{Gait-Based Gender Detection}

Table \ref{tab:gender-direct-combined} presents results for direct transfer (zero-shot) performance for gender estimation on CASIA-gender and FVG-gender. In this case, we compared different sizes of GaitFormer trained on DenseGait in two manners: i) only estimating attributes, without a constrastive objective (Attributes Only), and ii) estimating attributes and identity using a constrastive objective (MT). Similarly to the case of gait recognition, the Multi-Task networks consistently outperforms the other training regime. Moreover, the networks achieved reasonable performance in terms of F$_1$ score (76.18\% for CASIA-gender and 86.81\% for FVG-gender), considering that the networks were not exposed to any manually \mbox{annotated data}.

\begin{table}[hbt!]
\small
    \caption{GaitFormer direct transfer performance on gait-based gender estimation on CASIA-gender and FVG-gender. We highlight in \textbf{bold} the best overall results for each dataset.}
    \resizebox{\linewidth}{!}{
    \begin{tabular}{cccccccc}
     \toprule
        & & \multicolumn{3}{c}{\textbf{CASIA-Gender}} &  \multicolumn{3}{c}{\textbf{FVG-Gender}} \\
        \toprule
        \textbf{Size} & \textbf{Training} &  \textbf{Prec. }&  \textbf{Recall} &     \textbf{F\boldmath{$_1$}} & \textbf{Prec.} &  \textbf{Recall} &     \textbf{F\boldmath{$_1$}} \\
        \midrule
                SM &      Attributes &      94.54 &   62.46 &  72.10 & 85.87 &  84.9 &  85.10\\
                MD &      Attributes &      94.48 &   63.66 &  73.07  &    82.03 &  79.87 &  80.27\\
                XL &      Attributes &      94.59 &   62.43 &  72.08 &    84.36 &  84.43 &  84.26\\
                \midrule
                SM &       MT &      \textbf{94.84} &   63.61 &  73.00  &    86.47 &  86.34 &  86.33 \\
                MD &       MT &      94.71 &   61.50 &  71.31  &    86.96 &  \textbf{86.87} &  \textbf{86.81}\\
                XL &       MT &      94.72 &   \textbf{67.67} & \textbf{76.18} &    \textbf{87.06} &  86.21 &  86.38\\\bottomrule
        \end{tabular}
        }
    \label{tab:gender-direct-combined}
    \end{table}

Figure \ref{fig:gender-tunes} presents the performance under fine-tuning of GaitFormerXL on CASIA-gender and FVG-gender, in similar conditions to the recognition task. All networks are trained with a binary-crossentropy objective on the gender estimation task, without taking the person identity into account at training time. GaitFormerXL under Multi-Task training regime is consistently superior to a network initialized from random weights, achieving an F$_1$ score of 93.09\% on CASIA-gender and of 91.51\% on FVG-gender. The pretrained models significantly benefit from fine-tuning when small amounts of training data is available. Performance slightly increases with the availability of more training data.

{
\begin{figure*}[hbt!]
    \centering
    \includegraphics[width=0.98\linewidth]{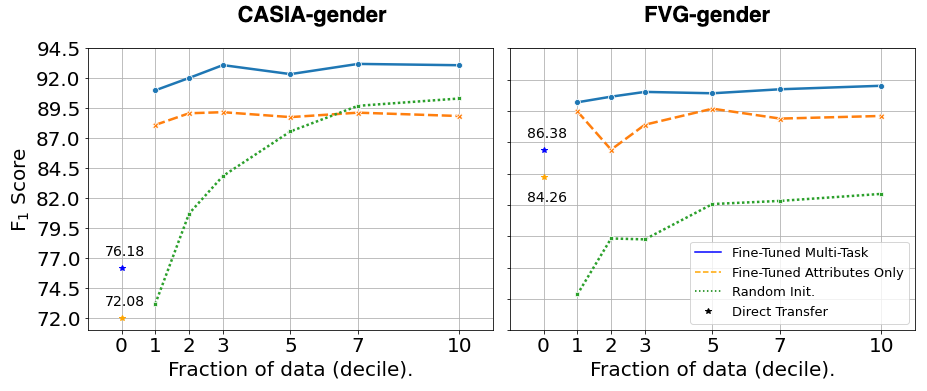}
    \caption{Fine-tuning results for GaitFormer on CASIA-gender and FVG-gender, trained on progressively larger samples of the datasets. Compared to a randomly initialized network, GaitFormer benefits significantly from fine-tuning in extremely low data regimes (e.g., 10\% of available annotated data). Compared to only pretraining on predicting attributes (Attributes Only), the Multi-Task network has consistently better performance across all fractions of the datasets.}
    \label{fig:gender-tunes}
\end{figure*}
}

\subsection{Gait-Based Pedestrian Attribute Identification}

For pedestrian attribute identification, we process a 10-h surveillance stream, corresponding to 10,733 tracklets, and use it for testing. For evaluation, we use the attribute pseudo-labels annotated automatically by the PAI ensemble. Figure \ref{fig:r2-pai} showcases R$^2$ score results for GaitFormerMD trained with a multi-task objective. This score is computed relative to the soft pseudo-labels estimated by the PAI ensemble. We emphasize that the model only uses movement information to estimate these labels, and has no information regarding appearance. 
Using a skeleton-based model for pedestrian attribute identification is useful in situations where the appearance of the person is unavailable (i.e., in privacy-critical scenarios). The model is effectively distilling external appearance into movement representations.

The model obtains good results in categories such as \textit{Gender}, \textit{AgeGroup}, \textit{BodyType} and \textit{Viewpoint}. The model is able to obtain better than average performance on categories such as \textit{Footwear}, and some types of clothing. However, some clothing categories have proven to be very difficult to model, especially \textit{LongCoat} and \textit{Trousers}. We hypothesize that such pieces of clothing negatively affect the accuracy of the pose estimation model, resulting in low quality extracted skeletons.

These are promising results which show that external appearance and movement are intrinsically linked together. This is evident in the more explicit relationship between, for example, footwear and gait, in which, intuitively, gait is severely affected by the walker's choice of shoes. Clothing, accessories, and actions while walking can be regarded as ``distractor'' attributes, which affect gait only temporarily. However, there are more subtle information cues which are present in gait, related to the developmental aspects of the person (e.g., gender, age, body composition, mental state etc). These attributes are more stable in time, and can provide insights into the internal workings of the walker. We posit that, in the future, works in gait analysis will tackle more rigorously the problem of estimating the internal state of the walker (i.e., personality/mental issues) through specialized datasets and methods.

\begin{figure*}[hbt!]
    \centering
    \includegraphics[width=0.96\linewidth]{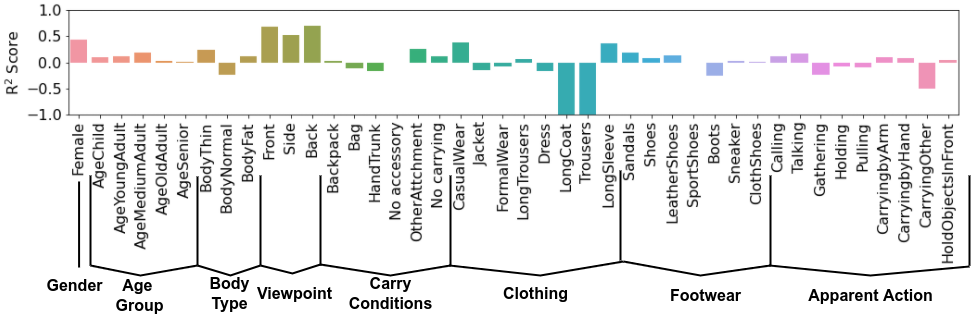}
    \caption{GaitFormerMD performance in terms of R$^2$ score. GaitFormerMD was trained with the multi-task objective. The model uses only movement information to predict attributes, and no information regarding the appearance of the individual.} %MDPI: 1. please confirm if italic is necessary? Author: No, removed. 2.  please confirm if different colors need explanations in image? Author: no explanations are    necessary.
    \label{fig:r2-pai}
\end{figure*}

\subsection{Inference Time}

Using transformer architectures for processing gait has other advantages besides a noticeable increase in downstream performance. Transformers have been shown to be more efficient in terms of inference time when compared to convolutional networks \cite{dosovitskiy2020image}. This effect is not directly correlated with the number of parameters, but is rather more influenced by the network structure \cite{9286182}. 

In Figure \ref{fig:inference-time}, we show a comparison between multiple sizes of GaitFormer, a plain transformer module minimally adapted for processing skeleton sequences, with the ST-GCN network, a popular architecture for skeleton action recognition \cite{yan2018spatial} and gait \mbox{analysis \cite{li2020jointsgait}}. We computed the inference time across multiple period lengths (from 12 frames to \mbox{96 frames}) to evaluate the scalability when processing shorter/longer sequences. For each period length, we run 100 experiments with a batch size of 512 and show the mean inference time in seconds, along with the standard deviation. All experiments were run on a NVIDIA RTX 3060 GPU. Even with comparable and exceeding number of parameters (ST-GCN from \cite{cosma2021wildgait} has 3.11M parameters), the transformer architecture clearly outperforms graph-convolutional models for processing gait sequences across multiple sequence lengths.

\begin{figure}[hbt!]
    \centering
    \includegraphics[width=0.9\linewidth]{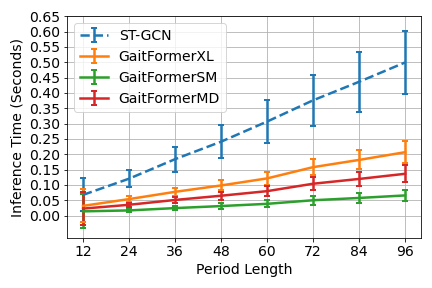}
    \caption{Inference times across processed walking duration length (period length) for ST-GCN and the various sizes of GaitFormer. We report the mean and stardard deviation across 100 runs, for each period length.}
    \label{fig:inference-time}
\end{figure}

\section{Conclusions}
In this work, we presented DenseGait, currently the largest dataset for pretraining gait analysis models, consisting of 217K anonymized skeleton sequences. Each skeleton sequence is  automatically annotated with 42 appearance attributes by making use of an ensemble of pretrained PAI networks. We make DenseGait available to the research community, under open credentialized access, to promote further advancement in the skeleton-based gait analysis field. We proposed GaitFormer, a transformer that is pretrained on DenseGait in a self-supervised and multi-task fashion. The model obtains 92.5\% accuracy on CASIA-B and 85.3\% accuracy on FVG, without processing any manually annotated data, achieving higher performance even compared to fully supervised methods. GaitFormer represents the first application of plain transformer encoders for skeleton-based gait analysis, without any hand-crafted architectural modifications. We explored pedestrian attribute identification based solely on movement, without utilizing appearance information. GaitFormer achieves good results in gender, age body type, and clothing attributes.

\section{Acknowledgements}
This work was partly supported by CRC Research Grant 2021, with funds from UEFISCDI in project CORNET (PN-III 1/2018) and by the Google IoT/Wearables Student Grants.

\bibliographystyle{ieee_fullname}
\bibliography{refs}

\end{document}